\documentclass[11pt,letterpaper]{berkeley}

\usepackage[authoryear,round]{natbib}
\usepackage[all]{hypcap}
\usepackage{inconsolata,placeins}
\usepackage{latexsym}
\usepackage[T1]{fontenc}
\usepackage[utf8]{inputenc}
\usepackage{microtype}
\usepackage{url}

\usepackage{amsmath,amssymb,amsfonts,amsthm}
\usepackage{bbm,dsfont,mathrsfs,mathtools,nicefrac}
\usepackage{graphicx}
\usepackage{booktabs,multirow,makecell,tabularx}
\usepackage{pdflscape}
\usepackage{algorithm}
\usepackage[noend]{algpseudocode}
\usepackage{epigraph,enumitem,float,listings,setspace,subcaption,wrapfig}
\usepackage{tikz}
\usetikzlibrary{arrows.meta,positioning}
\usepackage{xcolor}
\usepackage{xspace}
\usepackage{xparse}
\usepackage[normalem]{ulem}
\usepackage{pifont}
\usepackage{fvextra}
\usepackage[most,skins,theorems]{tcolorbox}
\tcbuselibrary{skins,breakable}
\usepackage[textsize=tiny]{todonotes}
\hypersetup{
    colorlinks=true,
    linkcolor=blue,
    urlcolor=blue,
    citecolor=blue,
    anchorcolor=blue
}
\usepackage[capitalize,noabbrev]{cleveref}
\usepackage{crossreftools}
\usepackage{titlesec}

\usepackage{amsmath,amsfonts,bm}

\def\eqref#1{equation~\ref{#1}}

\def\1{\bm{1}}

\DeclareMathAlphabet{\mathsfit}{\encodingdefault}{\sfdefault}{m}{sl}
\SetMathAlphabet{\mathsfit}{bold}{\encodingdefault}{\sfdefault}{bx}{n}

\definecolor{deepblue}{rgb}{0,0,0.5}
\definecolor{deepred}{rgb}{0.6,0,0}
\definecolor{deepgreen}{rgb}{0,0.5,0}
\definecolor{lightblue}{rgb}{0.22,0.45,0.70}
\definecolor{rliableolive}{HTML}{BBCC33}
\definecolor{rliableblue}{HTML}{77AADD}
\definecolor{rliablered}{HTML}{EE8866}

\newcommand\pythonstyle{\lstset{
basicstyle=\ttfamily\footnotesize,
language=Python,
morekeywords={self, clip, exp, mse_loss, uniform_sample, concatenate, logsumexp},
keywordstyle=\color{deepblue},
emph={MyClass,__init__},
emphstyle=\color{deepred},
stringstyle=\color{deepgreen},
frame=single,
showstringspaces=false
}}

\lstnewenvironment{python}[1][]
{
\pythonstyle
\lstset{#1}
}
{}

\newcommand\pythoninline[1]{{\pythonstyle\lstinline!#1!}}

\makeatletter
\def\mathcolor#1#{\@mathcolor{#1}}
\def\@mathcolor#1#2#3{%
  \protect\leavevmode
  \begingroup
    \color#1{#2}#3%
  \endgroup
}
\makeatother

\theoremstyle{plain}

\theoremstyle{definition}

\theoremstyle{remark}

\tcbset{
  aibox/.style={
    enhanced,
    breakable,
    width=\linewidth,
    top=7pt,
    bottom=5pt,
    left=6pt,
    right=6pt,
    boxsep=3pt,
    colback=blue!6!white,
    colframe=black,
    colbacktitle=black,
    fonttitle=\bfseries,
    attach boxed title to top left={yshift=-0.1in,xshift=0.15in},
    boxed title style={
      boxrule=0pt,
      colframe=white,
    },
  }
}

\NewDocumentCommand{\AIBoxStart}{m m}{%
  \IfNoValueTF{#2}{%
    \IfNoValueTF{#1}{%
      \begin{tcolorbox}[aibox]%
    }{%
      \begin{tcolorbox}[aibox,title={#1}]%
    }%
  }{%
    \IfNoValueTF{#1}{%
      \begin{tcolorbox}[aibox,title={#2}]%
    }{%
      \begin{tcolorbox}[aibox,title={#2},#1]%
    }%
  }%
}

\NewDocumentEnvironment{AIBox}{o g}
  {\AIBoxStart{#1}{#2}}
  {\end{tcolorbox}}

\NewDocumentEnvironment{AIbox}{o g}
  {\AIBoxStart{#1}{#2}}
  {\end{tcolorbox}}

\newtcolorbox{promptbox}[2][]{
listing only,
enhanced,
breakable,
colback=rliableolive!13!white,
colframe=black,
fontupper=\ttfamily,
title=#2,
#1}

\Crefformat{equation}{#2Eq.\;(#1)#3}
\Crefformat{figure}{#2Figure #1#3}
\Crefformat{assumption}{#2Assumption #1#3}
\Crefname{assumption}{Assumption}{Assumptions}

\newcommand{\bDiamond}{\mathbin{\Diamond}}

\newcommand{\methodname}{Librarian Agents}

\title{\methodname: Training LLMs to Curate Long-Context Workspaces that Anticipate Future Queries}

\reportnumber{} %

\author[1,$\bDiamond$]{Anikait Singh}
\author[1,$\bDiamond$]{Teresa Zhang}
\author[1,$\bDiamond$]{Yoonho Lee}
\author[1]{Roshen Sanjay Nair}
\author[2]{Aviral Kumar}
\author[1]{Chelsea Finn}

\affil[1]{Stanford University}
\affil[2]{Carnegie Mellon University}
\affil[$\bDiamond$]{Equal Contribution}
\correspondingauthor{anikait@stanford.edu.\\ \textbf{Project Website:} \url{https://librarian-agents.github.io/}, \textbf{Code}: \url{https://github.com/teresaz05/librarian-agents}}

\begin{abstract}
\vspace{-0.3cm}
Reasoning over long document contexts remains a challenge for large language models (LLMs). Standard retrieval methods are inherently reactive, bottlenecking agents with complex, cross-document reasoning at inference time, while precomputed structures like knowledge graphs often fall short because they rely on rigid heuristics. Instead, can we allow agents to perform \emph{anticipatory directed curation}: learning to proactively reorganize unstructured corpora prior to inference to efficiently serve diverse future queries. We build on a key asymmetry in modern agentic harnesses (e.g., Claude Code): while LLM agents are adept at navigating existing filesystems to solve an immediate task, they are rarely trained to construct reusable workspaces for future use. To instantiate this capability, we introduce \methodname{}, a framework in which a builder agent transforms raw documents into a hierarchical filesystem of synthetic files and reference links, which forms a workspace tailored for a downstream answering agent. We formulate workspace construction as a Markov decision process and optimize the builder policy via reinforcement learning to directly maximize downstream answering accuracy. Empirically, \methodname{} outperforms the strongest retrieval and precomputation baselines by 33 percentage points on BrowseComp-Plus, amortizes upfront construction costs across repeated queries, and generalizes zero-shot to unseen domains on BRIGHT with a 31-point average accuracy improvement.
\end{abstract}

\begin{document}

\maketitle

\section{Introduction}

Large language models (LLMs) are increasingly tasked with reasoning over extensive corpora of documents, often found in domains such enterprise knowledge bases, complex software repositories, and user chat histories, that exceed the context windows of today's LLMs. In these domains, the information required to complete a given task is rarely centralized; instead, it is dispersed and disconnected across many disparate sources. Thus, for each incoming request, agents repeatedly process and revisit the corpus from scratch. Can we instead enable systems to efficiently navigate these massive context spaces without incurring high, redundant computational costs each time a user poses a new query?

A standard approach to this challenge is retrieval-augmented generation (RAG)~\citep{Gao2023RAGSurvey}, in which a system responds to a query by searching raw documents, retrieving relevant passages, and synthesizing an answer. Recent work has strengthened this pipeline with multi-agent frameworks~\citep{Jiang2023FLARE,Asai2023SelfRAG,Yan2024CRAG,Edge2024GraphRAG,Singh2025AgenticRAGSurvey}. However, these methods remain largely reactive: each query triggers a new round of search, requiring the model to repeatedly rediscover useful cross-document relationships at inference time. Other approaches pre-structure the search space before inference, for example, as a knowledge graph~\citep{peng2025graph}. These methods can improve retrieval, but they typically commit to a rigid set of abstractions such as extracting named entites~\citep{keraghel2024recentadvancesnamedentity}. Such a structure is useful when future queries precisely align with the index, but it is less natural when queries require new task-specific intermediate artifacts. A common underlying theme of these approaches is that they impose an inductive bias of structure, which a downstream agent can utilize when answering a given query. We ask can we learn to anticipate this structure for a new set of context that we encounter, curating these documents in a form that that an answering agents can readily use. We propose to directly train an agent to reorganize a corpus into an anticipatory \textit{answer-facing workspace} so that future LLM agents can efficiently find the right evidence for a query?

\begin{figure*}[t]
    \centering
    \includegraphics[width=0.9\linewidth]{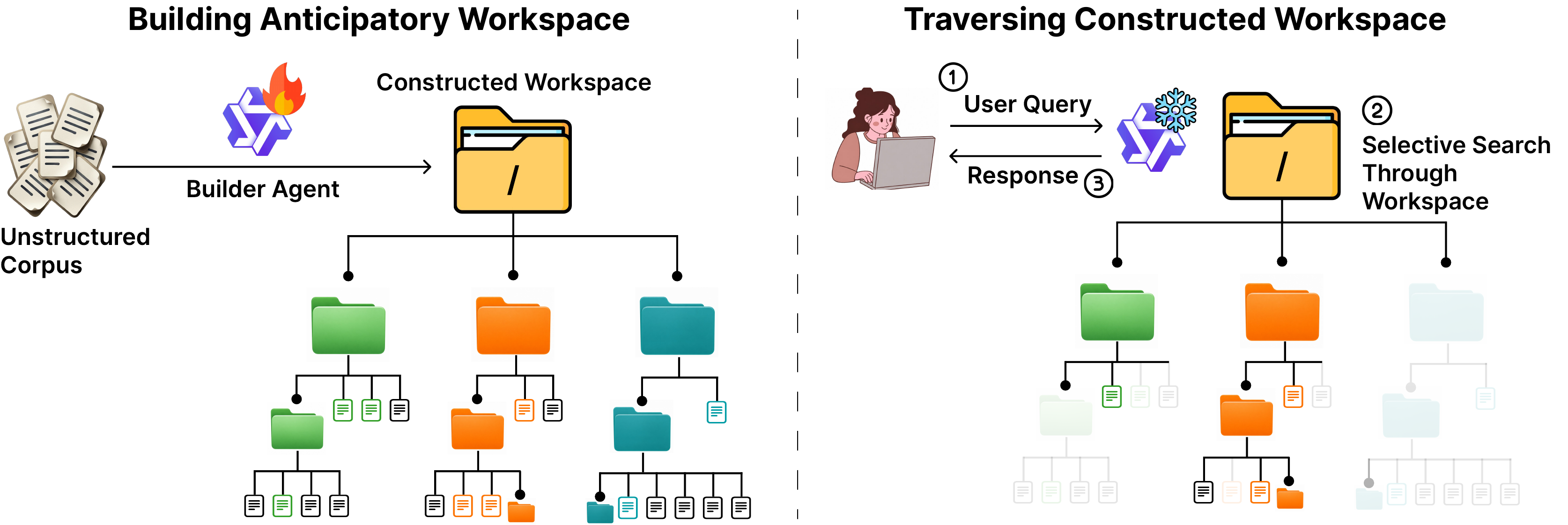}
     \caption{\footnotesize \textbf{Overview of \methodname{}.} (Left) We train a builder agent to construct a workspace consisting of synthetic files and directories that \emph{anticipates} future queries for a downstream answering agent. (Right) This workspace is built once and repeatedly used to answer different queries.}
    \label{fig:teaser}
    \vspace{-0.2cm}
\end{figure*}

Motivated by approaches that shift computation across different aspects of the model training and deployment process~\citep{Lin2025SleepTimeCompute, Wu2026ReasoningCache, suzgun2025dynamiccheatsheettesttimelearning, Zhang2025RLM}, we study \emph{anticipatory workspace construction}: the problem of augmenting and reorganizing a corpus \textit{before} future queries arrive. The goal is to shift part of the costs from query time to a precomputation phase, thereby amortizing these costs across many downstream queries. To instantiate this setting, we introduce \methodname{}, a framework built on a key asymmetry in current agentic systems: LLM agents are effective at navigating existing hierarchical filesystems, but are rarely trained to \textit{proactively construct such workspaces for future use}. \methodname{} addresses this gap by tasking a \textbf{builder agent} with transforming source documents into a multi-level workspace with references to the original corpus. An frozen \textbf{answering agent} then navigates this workspace to answer an individual user query. We formalize the builder agent's operations as a Markov Decision Process (MDP) and directly optimize the workspace construction policy for downstream answering performance by posing it as a reinforcement learning (RL) problem on this MDP.

We demonstrate the efficacy of \methodname{} on BrowseComp-Plus, \methodname{} improves accuracy by an average of 33 percentage points over the best performing baseline. Furthermore, for BRIGHT, a domain which our builder agent is untrained on, gains persist over baselines with an average 31 point improvement over baselines. 

In summary, our key contributions are:
\begin{itemize}
    \item We formulate anticipatory workspace construction as a learnable model behavior: agents should materialize useful context as an answer-facing workspace \textit{before} receiving user queries.
    \item We propose \methodname{}, a framework where a builder agent synthesizes and organizes memory artifacts into a filesystem-based workspace for future queries.
    \item We formulate filesystem-based workspace construction as an MDP and optimize the builder agent via reinforcement learning to improve downstream query-answering performance.
    \item We evaluate \methodname{} against retrieval and pre-computation baselines on BrowseComp-Plus and test zero-shot transfer on question-answering tasks derived from BRIGHT.
\end{itemize}
\section{\methodname{}: Agent Curated Filesystems as Anticipatory Workspaces}
\label{sec:methods}
We introduce \methodname{}, a framework for training an agent to construct an \emph{anticipatory workspace}: a filesystem-based memory representation built before future queries arrive. Document-grounded agents can waste compute when each new query forces them to rediscover the same information across the raw corpus. In contrast, can we shift part of this computation by constructing a reusable memory in the form of an anticipatory workspace? We instantiate this by have an gent curate an augmented filesystem that exposes a long-context structure through operations that these agents already use well: using tools to navigate filesystems~\citep{yang2024sweagent}. The builder converts a flat collection of documents into a multi-level workspace whose synthetic files reduce search friction while preserving reference paths to the original corpus. 

We first define the document corpus, filesystem, and augmentation operations that transform a flat collection of raw documents into an answer-facing workspace. We then articulate desiderata of an effective anticipatory memory workspace, which includes coverage with efficient navigation, fallback to raw evidence, and reuse across queries. Finally, we cast workspace construction as a finite-horizon Markov decision process in which the librarian model edits the filesystem to maximize the answering agent's expected performance on future queries, subject to a construction-cost penalty.

\subsection{Problem Setup.}

We study long-context reasoning over a corpus of raw documents, $\mathcal{D} = \{d_1, d_2, \ldots, d_n\}$, where the system must answer queries $q \sim \mathcal{Q}$. This setting matches retrieval-augmented generation (RAG) systems, which typically answer each query by searching over a static retrieval index over $\mathcal{D}$ and then synthesizing a response.

We represent the agent's external memory as a filesystem-based \emph{workspace} $\mathcal{F}$: a set of files and directories together with metadata specifying each node's type and parent.
The initial workspace $\mathcal{F}_0$ contains the raw documents in $\mathcal{D}$ as files under the root. 
An \textbf{augmentation} $\mathcal{A}$ is an edit operation that modifies a workspace, for example by creating, rewriting, deleting, or moving files and folders. At iteration $i$, an augmentation $\mathcal{A}_i$ transforms the previous workspace $\mathcal{F}_{i-1}$ into $\mathcal{F}_i$. Applying a sequence of augmentations for horizon $H$ produces a final workspace $\mathcal{F}_H$.

System performance is measured by a scoring function $\text{Acc}(\tilde{y}, y^*)$ that evaluates the correctness of the generated response $\tilde{y}$ against the ground-truth target $y^*$. Memory construction also incurs a cost, denoted $\text{Cost}(A_i)$, such as the number of tokens generated or modified by $\mathcal{A}_i$.

\begin{AIbox}[Overview of Problem Setup]
\textit{Anticipatory workspace construction} is the problem of transforming a large corpus into a structured filesystem-based workspace \uline{before} downstream queries are observed.
\end{AIbox}

\subsection{Desiderata for Anticipatory Workspaces.} 
A useful memory should do more than ``remember'' a corpus. It should reshape the corpus into an \textit{answer-facing workspace} that helps future agents quickly find relevant evidence while preserving a path back to the raw source documents. As a running example, consider an enterprise corpus containing design docs, issue threads, launch notes, and customer reports for the same product. One future query may ask for a low-level implementation detail; another may ask how a user-visible behavior changed across releases; a third may ask for the exact source that supports a disputed claim. We use these examples to motivate three key desiderata.

\textbf{Coverage with Efficient Navigation.}
The memory should cover many plausible queries while reducing the need to search the raw corpus. In the enterprise example, the builder might create a release timeline, a component-level design index, and customer-facing behavior summaries, each linking the same underlying documents from a different angle. These artifacts help only if the answering agent can reach them through short, legible paths, such as directories named by abstraction level. Thus, the builder should synthesize cross-document artifacts that cover likely regions of the query distribution $\mathcal{Q}$ and place them where an agent can discover them with shallow traversal.

\textbf{Faithful Fallback.}
No finite memory can anticipate \textit{every} query, and synthesized artifacts can omit details that are critical in hindsight. The workspace must therefore preserve reference links and raw source paths so the answering agent can verify claims or bypass the synthesized layer when needed. In the enterprise example, a rare query about a single customer may not appear in the high-level summary, but the agent should still be able to read any existing issue thread. As an extreme example, such a fallback is important for needle-in-a-haystack queries where the answer depends on a localized, very specific fact that is unlikely to survive compression~\citep{wang2024needlemultimodalhaystack}.

\textbf{Amortized Reuse.}
The builder's construction cost should pay off across many downstream queries. For example, a compact release timeline written by the builder agent may help answer many questions about regressions, roadmap changes, customer messaging, and dependency updates. In contrast, a one-off answer draft may surface similar information but help only a single request. This creates a tradeoff between the construction cost $\sum_{i=1}^H \mathrm{Cost}(\mathcal{A}_i)$ and the downstream utility.

\begin{AIbox}[Overview of Anticipatory Workspace Desiderata]
An effective anticipatory workspace should make likely queries easy to answer, preserve fallback paths to raw documents, and pay off across many downstream queries.
\end{AIbox}

\subsection{Workspace MDP Formalization.}
\label{subsec:method-mdp}

In \methodname{}, anticipatory workspace construction proceeds in two phases: an upfront synthesis phase and a downstream answering phase. The synthesis phase is analogous to the offline construction of a retrieval index or knowledge graph, but instead produces a flexible filesystem-based workspace. To instantiate this idea, we define a \textbf{builder agent} $\pi_{\text{build}}(\mathcal{A} \mid \mathcal{F})$, which proposes augmentations given the current workspace, and an \textbf{answering agent} $\pi_{\text{ans}}(\tilde{y} \mid q, \mathcal{F})$, which generates an answer $\tilde{y}$ for a new query $q$ using the constructed workspace.

\begin{figure*}[h]
    \centering
    \includegraphics[width=\linewidth]{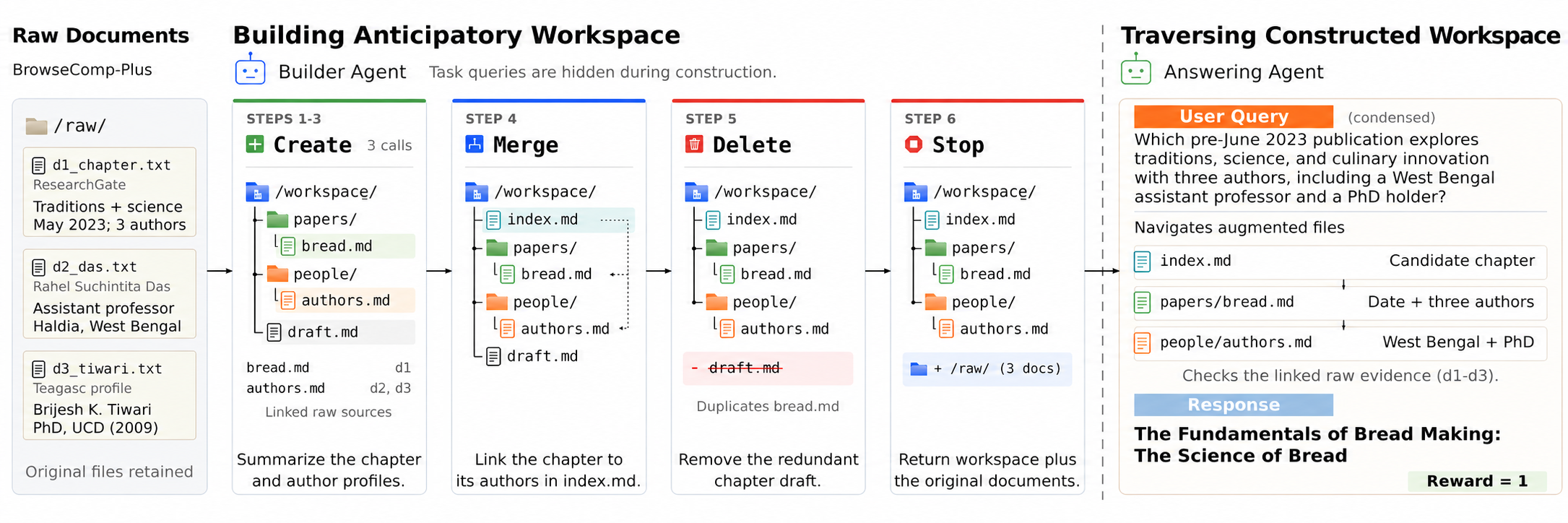}
    \caption{
        \footnotesize \textbf{MDP Definition of \methodname{}.} Starting from a raw list of documents, a builder agent can create a workspace with four actions: \texttt{Create}, \texttt{Merge}, \texttt{Delete}, and \texttt{Stop}. An answering agent navigates the resulting workspace through file-listing and file-reading actions, produces a response, and receives a task-specific correctness score. Here, we depict an augmented workspace from the BrowseComp-Plus dataset. 
    }
    \label{fig:mdp-browsecomp}
\end{figure*}

We cast the builder agent's sequence of filesystem operations as a finite-horizon Markov Decision Process (MDP), allowing memory construction to be optimized as a policy-learning problem.

\textbf{State, Action, and Transition Dynamics.}
At each timestep $t$, the state is the current filesystem $\mathcal{F}_t$, consisting of the original raw documents $\mathcal{D}$ together with the synthetic files, folders, and reference links created so far. The builder agent selects an augmentation $\mathcal{A}_t \sim \pi_{\text{build}}(\cdot \mid \mathcal{F}_t)$, which is applied to the filesystem to produce $\mathcal{F}_{t+1}$. We constrain the builder to the following discrete action types that can each be applied to different files:
\begin{itemize}
    \item \textsc{CreateFile}: Creates a new synthetic file by consolidating information from a subset of raw documents and/or existing synthetic files. The file has reference links to the set of sources used to construct it.
    \item \textsc{MergeFiles}: Aggregates a set of related synthetic files into a higher-level summary or directory-like artifact. This action introduces hierarchy, linking to child files.
    \item \textsc{Delete}: Removes redundant/unhelpful synthetic files from the active answer-facing workspace. This operation never deletes raw documents in $\mathcal{D}$, preserving fallback access.
    \item \textsc{Stop}: Terminates memory construction, returning the current workspace $\mathcal{F}_t$ as the final memory artifact to use for the downstream answering agent.
\end{itemize}
The action types here is intentionally constrained to reduce the horizon for optimization. In principle, this interface can be generalized to a broader set of atomic filesystem commands (e.g., \texttt{touch}, \texttt{mkdir}, \texttt{mv}, \texttt{rm}, etc). We describe the instantiation of these actions in the filesystem in Appendix~\ref{app:tool-use}.

\textbf{Reward Function.}
The reward should incentivize the builder to construct a memory artifact that improves both downstream answer accuracy and navigation efficiency over the query distribution~$\mathcal{Q}$. We first define the expected answering success of a filesystem~$\mathcal{F}$ as
\begin{equation*}
  S(\mathcal{F}) \coloneqq
  \mathbb{E}_{q \sim \mathcal{Q}}\,
  \mathbb{E}_{\tilde{y} \sim \pi_{\mathrm{ans}}(\cdot \mid q, \mathcal{F})}
  \bigl[ \mathrm{Acc}\bigl(\tilde{y}, y^{*}(q)\bigr) \bigr],
\end{equation*}
where $y^{*}(q)$ is the ground-truth answer to query~$q$ and $\mathrm{Acc}(\tilde{y}, y^{*}(q)) \in \{0, 1\}$ indicates whether the answer~$\tilde{y}$ is correct. To account for navigation efficiency, we define the downstream utility
\begin{equation}
  U(\mathcal{F}) \coloneqq
  S(\mathcal{F})
  - \lambda_{\mathrm{tok}} \, C_{\mathrm{tok}}(\mathcal{F})
  - \lambda_{\mathrm{read}} \, C_{\mathrm{read}}(\mathcal{F}),
  \label{eq:utility}
\end{equation}
where $C_{\mathrm{tok}}$ and $C_{\mathrm{read}}$ denote the expected normalized synthetic-memory token cost and file-read cost, respectively. Before averaging over queries and answering rollouts, each rollout's cost is multiplied by its correctness indicator. Navigation penalties therefore apply only to correct answers: an incorrect answer receives zero utility regardless of its navigation cost. Setting $\lambda_{\mathrm{tok}} = \lambda_{\mathrm{read}} = 0$ recovers the accuracy-only utility $U(\mathcal{F}) = S(\mathcal{F})$.

The sparse construction objective scores the final filesystem while penalizing the total cost of building it:
\begin{equation}
  r^{\mathrm{sparse}} \coloneqq
  U(\mathcal{F}_H)
  - \alpha \sum_{t=1}^{H} \mathrm{Cost}_{\mathrm{build}}(\mathcal{A}_t),
  \label{eq:sparse}
\end{equation}
where $\alpha \geq 0$ weights construction-time expenditure. Although this objective captures the desired outcome, it provides only coarse credit assignment over a potentially long sequence of edits. An ideal dense reward instead credits each augmentation with its marginal contribution to downstream utility, net of its construction cost:
\begin{equation}
  r_t^{\mathrm{dense}} \coloneqq
  U(\mathcal{F}_t) - U(\mathcal{F}_{t-1})
  - \alpha \, \mathrm{Cost}_{\mathrm{build}}(\mathcal{A}_t).
  \label{eq:dense-reward}
\end{equation}
Under undiscounted returns ($\gamma = 1$), these rewards telescope:
\begin{equation}
  \sum_{t=1}^{H} r_t^{\mathrm{dense}}
  = r^{\mathrm{sparse}} - U(\mathcal{F}_0).
  \label{eq:telescoping}
\end{equation}
Because $U(\mathcal{F}_0)$ is fixed for a given initial filesystem, maximizing the cumulative dense reward is equivalent to maximizing the sparse objective, following potential-based reward shaping~\citep{ng1999policy}. Advantages are computed at each step, following multi-turn GRPO implementations~\citep{baronio2025kevinmultiturnrlgenerating}, thereby allowing earlier builder turns to receive a denser learning signal.

\begin{figure}[t]
    \centering
    \includegraphics[width=0.9\linewidth]{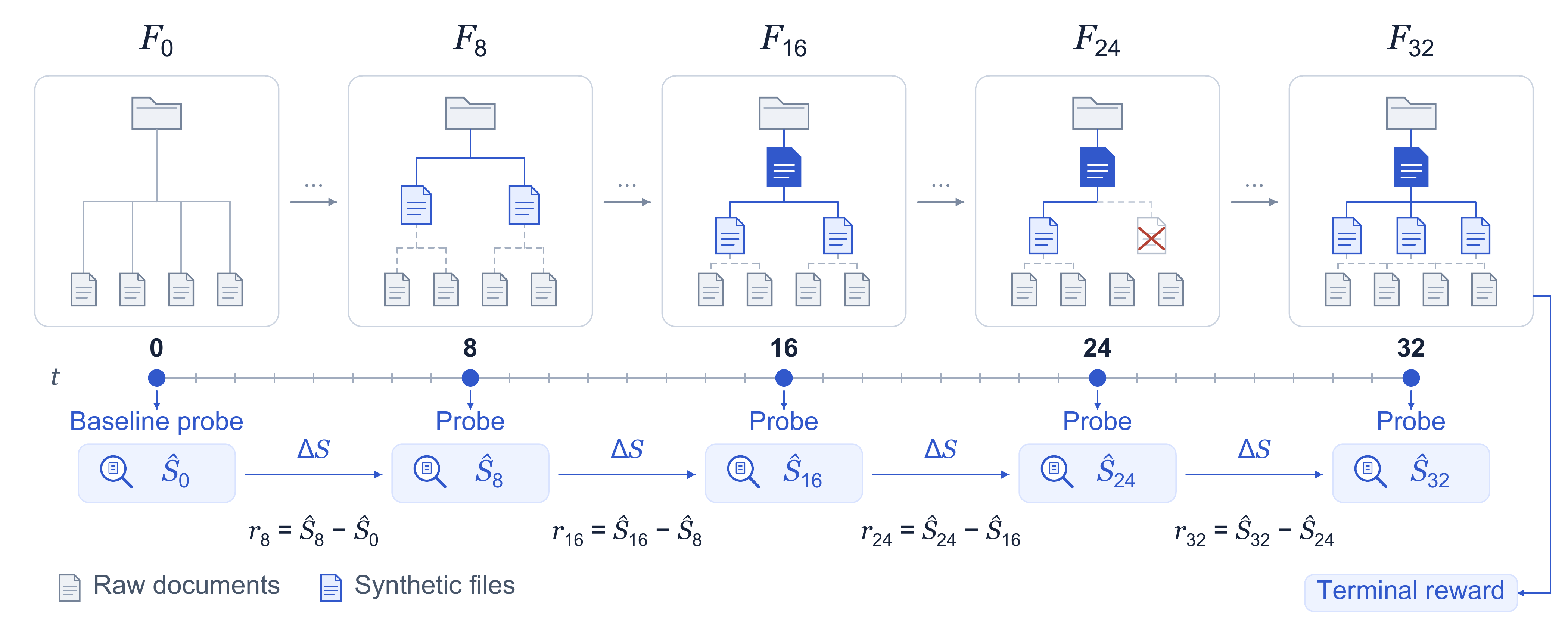}
    \caption{\footnotesize \textbf{Dense Rewards via Probes.} We train the builder agent with RL to directly maximize performance of answering agent. To densify the reward, the builder agent operates over workspace augmentations and is rewarded for change in accuracy. Probes to the answering agent are done at regular intervals (8 turns) up to the maximum horizon $H$.}
    \label{fig:probes}
    \vspace{-0.4cm}
\end{figure}

Evaluating downstream utility after every edit can be expensive, so we train with a practical surrogate objective. We probe answerability every eight turns as seen in Figure~\ref{fig:probes} and combine changes between consecutive probes with a terminal downstream-answer reward. In the primary experiment, we use $\gamma=1$ and optimize answerability and terminal correctness without explicit navigation-cost, filesystem-structure, or construction-cost shaping; equivalently, $\lambda_{\mathrm{tok}}=\lambda_{\mathrm{read}}=\alpha=0$. During training, all downstream quantities are estimated on minibatches of training queries that are hidden from the builder; evaluation queries are never used to construct a filesystem or select the reported checkpoint. Appendix~\ref{app:hyperparam} provides the complete configuration.

\begin{AIbox}[Overview of MDP Formulation]
We formulate memory construction as an MDP in which periodic answerability changes and terminal answer correctness provide credit for useful filesystem edits; the general objective can additionally incorporate navigation and construction costs.
\end{AIbox}

\subsection{Workspace Augmentations via RL.}
\label{subsec:method-rl}
The MDP formulation above enables direct optimization of the workspace construction process with reinforcement learning. 
In each rollout, the builder iteratively edits the workspace for up to $H$ steps, producing a trajectory
\begin{equation}
    \tau = (\mathcal{F}_0, \mathcal{A}_1, \mathcal{F}_1, \ldots, \mathcal{A}_T, \mathcal{F}_T),
\end{equation}
where $T \leq H$ is determined either by the horizon or the \textsc{Stop} action. 

The resulting workspace is then evaluated by the answering agent on a minibatch of queries sampled from $\mathcal{Q}$. These queries are used only for reward estimation and are \textbf{not} provided to the builder during construction. In the primary configuration, the builder receives periodic answerability-delta rewards together with the terminal answer score. This hybrid signal approximates the dense objective while avoiding an expensive answering evaluation after every workspace edit. This training signal encourages the builder to discover workspace edits that are not merely locally coherent but also useful for improving downstream question answering. When $\alpha>0$, the objective additionally trades these improvements against construction cost. We optimize the builder policy with GRPO~\citep{shao2024deepseekmath}, a policy-gradient method used for RL-based fine-tuning of language models.

By default, we treat the answering agent $\pi_{\text{ans}}$ as fixed during builder optimization. This isolates the learning problem: the builder learns to construct memories that are useful to the answering agent, which may be a language model or another long-context reasoning system such as RLM~\citep{Zhang2025RLM}. We can also consider an alternative joint-training variant in which the answering agent is adapted using the same downstream correctness signal. In this setting, optimization can alternate between updating the builder and updating the answerer via a cooperative multi-agent RL framework, following methods such as RLAD~\citep{qu2025rladtrainingllmsdiscover}. The full optimization process is outlined in Algorithm~\ref{alg:memory-construction}.

\begin{AIbox}[Overview of RL Optimization]
We optimize the builder as an LLM policy over filesystem edits. Each rollout constructs a memory, evaluates it with a fixed answering agent, and updates the builder with GRPO using periodic answerability changes and terminal answer correctness.
\end{AIbox}

Having formulated anticipatory workspace construction as an MDP in Section~\ref{sec:methods}, we evaluate whether \methodname{} improves the reasoning performance and inference efficiency of downstream answering agents on long-context document reasoning tasks. We compare \methodname{} with prior reactive retrieval and pre-computation methods to address four questions: \textbf{(1)} Does \methodname{} improve in-distribution long-context reasoning performance? \textbf{(2)} What amortization benefits do \methodname{} provide when serving multiple queries over a shared document set? \textbf{(3)} Does the builder agent’s anticipatory construction capability generalize zero-shot to out-of-distribution retrieval settings? \textbf{(4)} Qualitatively, how do the builder and answering agents behave during the construction and answering phases, and do these behaviors reflect the intended anticipatory capability? We examine each question in turn below.

\textbf{Task Definition.} The main domain we consider for long-document reasoning is BrowseComp-Plus~\citep{chen2025browsecomp}, a curated corpus of deep-research web tasks. We split this dataset into two sets for training and evaluation: 680 tasks for training and 50 held-out tasks for all evaluations (sampled at random). For training $\pi_\text{build}$, each task consists of an independent set of raw documents and a downstream query to be answered based on the set of documents. We additionally evaluate the performance of \methodname{} on the BRIGHT Benchmark~\citep{su2025brightrealisticchallengingbenchmark}, which comprises tasks from diverse domains (StackExchange, LeetCode, and math competitions), to assess the builder agent's efficacy on unseen distributions.

\noindent \textbf{Comparisons and baselines.}
We evaluate our approach against several representative baselines. First, we compare with standard reactive retrieval methods: an \emph{Embedding} with optional \emph{Reranker} baseline~\citep{zhang2025qwen3} and \emph{Late Interaction Embedding}~\citep{chaffin2025pylateflexibletrainingretrieval}, which represent conventional retrieve-and-read RAG systems that process each query from scratch. Second, we compare with \emph{Graph RAG} \citep{peng2025graph}, a knowledge graph baseline that uses precomputation. Third, we include \emph{Agentic RAG} \citep{Singh2025AgenticRAGSurvey}, which represents multi-agent approaches that dynamically interleave retrieval and reasoning at query time.  Next, we compare autocompaction methods such as those that periodically summarize the computation performed, such as RC \citep{Wu2026ReasoningCache}. These baselines cover both ends of the design space, with some doing all cross-document reasoning per query and others doing it once up front. We also include an ablation of our own framework: a \emph{builder agent without RL fine-tuning}. This isolates the contribution of our MDP-based policy optimization. For all methods, the downstream answering agents, Qwen3.5-4B and Gemini 3.1 Flash lite, are kept frozen so that performance differences reflect only the quality of the resulting memory artifacts. More details on the baselines and hyperparameters are available in Appendix~\ref{app:hyperparam}.

\begin{figure}[t]
    \centering
    \includegraphics[width=0.9\linewidth]{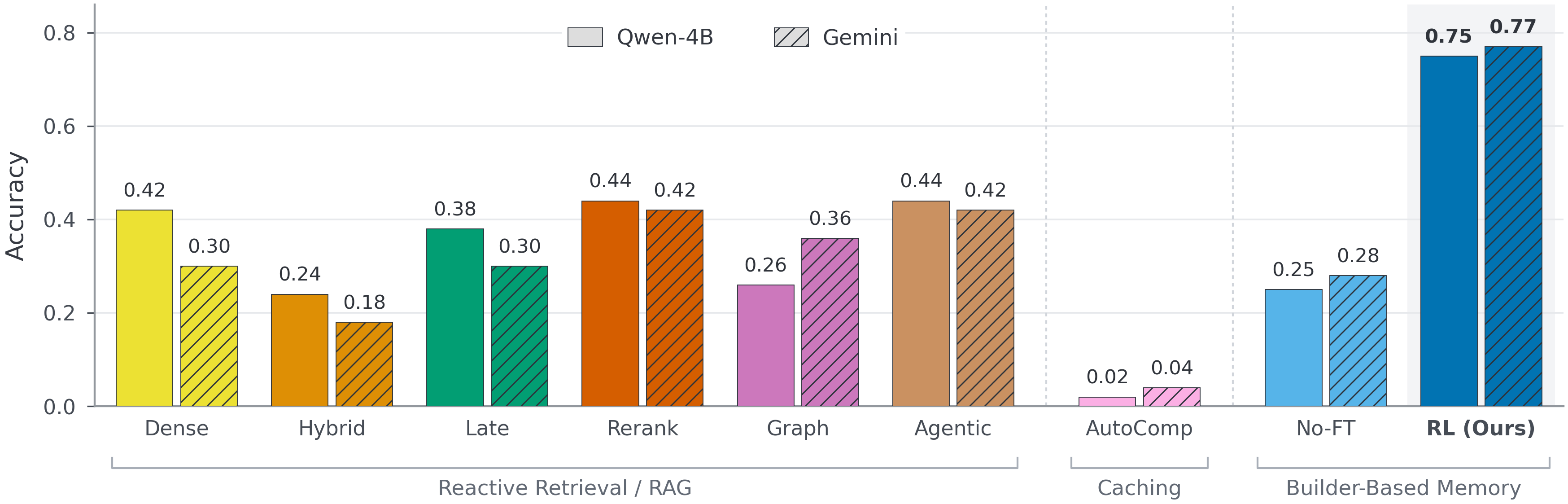}
    \caption{
        \footnotesize
        \textbf{Results on BrowseComp-Plus.}
        We compare \methodname{} against RAG, Caching and Builder-Based Memory baselines using two answering agent models (Qwen 3.5 4B and Gemini 3.1 Flash Lite). We find that \methodname{} significantly outperforms all methods for both answering agents. 
    }
    \label{fig:browsecomp_plus_results}
\end{figure}

\noindent \textbf{(1) In Distribution Performance of \methodname{} on BrowseComp-Plus.}
We evaluate the performance of \methodname{} on the BrowseComp-Plus domain in Figure~\ref{fig:browsecomp_plus_results}. Here we find that \methodname{} with RL fine-tuning significantly outperforms all reactive and pre-computation baselines. When using the frozen \textsc{Qwen3.5-4B} answering agent, \methodname{} achieves an accuracy of 0.75, representing a substantial improvement over the Embedding (0.42) and Reranker (0.44) baselines. A similar trend is observed with \textsc{Gemini 3.1 Flash Lite}, where \methodname{} reaches 0.77, compared with baseline scores of 0.30 and 0.42. Crucially, the ablation without RL fine-tuning (No FT) additionally underperforms below \methodname{}, highlights that policy optimization with RL is essential for teaching the builder agent to structure memories effectively for downstream consumption, a task the base model was not trained for.

\begin{table*}[t]
    \centering
    \scriptsize
    \setlength{\tabcolsep}{4pt}
    \renewcommand{\arraystretch}{1.08}

    \begin{minipage}[t]{0.48\linewidth}
        \vspace{0pt}
        \caption{
            \footnotesize
            \textbf{Construction Cost, BrowseComp-Plus.}
            We compare the accuracy and cost of \methodname{} with indexing based baselines in tokens.
        }
        \centering
        \begin{tabularx}{\linewidth}{@{}Xcc@{}}
            \toprule
            \textbf{Memory / Retrieval}
                & \textbf{Acc.}
                & \shortstack{\textbf{Construction Cost}\\
                              \textbf{(tokens)}} \\
            \midrule
            GraphRAG
                & 0.36 & 11,923 \\
            AutoCompaction
                & 0.04 & 501,637 \\
            Builder (No FT)
                & 0.28 & 127,683 \\
            \midrule
            \textbf{\methodname{}} (ours)
                & \textbf{0.77} & \textbf{178,926} \\
            \bottomrule
        \end{tabularx}
        \label{tab:browsecomp-plus-amortization}
    \end{minipage}\hfill
    \begin{minipage}[t]{0.48\linewidth}
        \vspace{0pt}
        \caption{
            \footnotesize
            \textbf{Performance on amortized BrowseComp-Plus
            (base vs.\ fine-tuned).}
            The fine-tuned builder improves accuracy from 0.24 to
            0.64 on the amortized variant of browsecomp. The resulting answering agent reads 26.9 files and 9,346 tokens per query on average.
        }
        \centering
        \begin{tabularx}{\linewidth}{@{}Xccc@{}}
            \toprule
            \textbf{Model}
                & \textbf{Accuracy}
                & \shortstack{\textbf{Answer tokens}\\
                              \textbf{per query}}
                & \shortstack{\textbf{Files}\\
                              \textbf{per query}} \\
            \midrule
            Qwen 3.5 4B Builder
                & \shortstack{0.242\\\(\pm\) 0.036} & \shortstack{1,238\\\(\pm\) 155} & \shortstack{2.72\\\(\pm\) 0.35} \\
            \textbf{Ours}
                & \shortstack{\textbf{0.637}\\\textbf{\(\pm\) 0.029}} & \shortstack{9,346\\\(\pm\) 596} & \shortstack{26.90\\\(\pm\) 1.70} \\
            \bottomrule
        \end{tabularx}
        \label{tab:browsecomp_amortized_results}
    \end{minipage}
\end{table*}

\noindent \textbf{(2) Memory reuse and cost per correct answer.}
Filesystem construction is a one-time cost that helps all queries using the same document set.
We therefore measure the \emph{amortized} cost in a dataset we extend so that each document set serves multiple queries.
As shown in Table~\ref{tab:browsecomp-plus-amortization}, \methodname{} incurs an upfront construction cost of 179K tokens and achieves an accuracy of 0.77, compared with 0.36 and 0.04, respectively. To examine how this cost is distributed when a filesystem is reused, we evaluate 443 frozen synthetic queries across 50 document sets, with 7--10 queries per document set, generated by Gemini 3.1 Flash Lite few-shot prompted with sample DeepDive~\citep{lu2025deepdive} questions. On these queries, the same evaluation filesystems yield an accuracy of 0.64 versus 0.24 for the base builder at an answer-time cost of 9.3K tokens per query. At a hypothetical reuse load of 20 queries per document set, spreading the reported construction cost across those queries gives a projected total of roughly 18K tokens per query (Table~\ref{tab:browsecomp_amortized_results}). Because the answering agent reads a richer filesystem, raw per-query token counts remain higher than with the base builder. After normalizing for accuracy, \methodname{} uses 33.8\% fewer recorded tokens per expected correct answer at ten questions per corpus (RL/Base ratio 0.662, 95\% CI [0.504, 0.836]) as shown in Appendix~\ref{app:browsecomp-break-even}.

\begin{table*}[t]
    \centering
    \caption{
        \footnotesize
        \textbf{Zero-Shot Transfer to BRIGHT.}
        The builder agent, trained only on DeepResearch tasks
        (BrowseComp) is evaluated on BRIGHT without task-specific training, testing whether \methodname{} learns a transferable capability of anticipation across different query and document sources. \methodname{} outperforms baselines on 3 sub-tasks within BRIGHT (AOPS, Biology, and Stack Overflow) achieving an average a pass@1 accuracy of 52\% and pass@3 accuracy of 67\% with the \textsc{Qwen3.5-4B} model.
    }
    \begingroup
    \scriptsize
    \setlength{\tabcolsep}{2.5pt}
    \renewcommand{\arraystretch}{1.18}
    \begin{tabular*}{\textwidth}{@{\extracolsep{\fill}}l*{8}{c}@{}}
        \toprule
        & \multicolumn{2}{c}{\textbf{AOPS}}
        & \multicolumn{2}{c}{\textbf{Biology}}
        & \multicolumn{2}{c}{\textbf{Stack Overflow}}
        & \multicolumn{2}{c}{\textbf{Average}} \\
        \cmidrule(lr){2-3}\cmidrule(lr){4-5}\cmidrule(lr){6-7}\cmidrule(l){8-9}
        \textbf{Method} & pass@1 & pass@3 & pass@1 & pass@3 & pass@1 & pass@3 & pass@1 & pass@3 \\
        \midrule
        BM25 & 30.5 \(\pm\) 5.9 & 37.0 \(\pm\) 6.8 & 17.5 \(\pm\) 4.6 & 26.0 \(\pm\) 6.0 & 9.0 \(\pm\) 3.3 & 15.5 \(\pm\) 4.8 & 19.0 \(\pm\) 2.7 & 26.2 \(\pm\) 3.4 \\
        Dense & 31.5 \(\pm\) 6.0 & 38.0 \(\pm\) 6.9 & 19.5 \(\pm\) 4.8 & 29.0 \(\pm\) 6.1 & 11.5 \(\pm\) 3.9 & 17.0 \(\pm\) 5.2 & 20.8 \(\pm\) 2.9 & 28.0 \(\pm\) 3.5 \\
        Hybrid & 33.0 \(\pm\) 6.4 & 37.0 \(\pm\) 6.8 & 14.0 \(\pm\) 4.3 & 20.5 \(\pm\) 5.6 & 7.0 \(\pm\) 2.4 & 15.5 \(\pm\) 4.8 & 18.0 \(\pm\) 2.7 & 24.3 \(\pm\) 3.3 \\
        Rerank & 32.0 \(\pm\) 6.1 & 37.5 \(\pm\) 6.9 & 14.0 \(\pm\) 4.1 & 23.0 \(\pm\) 5.6 & 6.0 \(\pm\) 2.6 & 11.0 \(\pm\) 4.3 & 17.3 \(\pm\) 2.6 & 23.8 \(\pm\) 3.3 \\
        Agentic & 28.5 \(\pm\) 5.6 & 37.0 \(\pm\) 6.8 & 17.0 \(\pm\) 4.5 & 25.5 \(\pm\) 5.9 & 6.5 \(\pm\) 2.5 & 14.5 \(\pm\) 4.5 & 17.3 \(\pm\) 2.5 & 25.7 \(\pm\) 3.3 \\
        Late interaction & 28.5 \(\pm\) 5.5 & 37.5 \(\pm\) 6.9 & 18.0 \(\pm\) 4.9 & 24.5 \(\pm\) 6.0 & 7.0 \(\pm\) 2.5 & 14.5 \(\pm\) 4.8 & 17.8 \(\pm\) 2.6 & 25.5 \(\pm\) 3.4 \\
        Graph & 12.5 \(\pm\) 4.4 & 15.5 \(\pm\) 5.1 & 5.0 \(\pm\) 2.9 & 6.0 \(\pm\) 3.4 & 1.5 \(\pm\) 1.1 & 3.5 \(\pm\) 2.5 & 6.3 \(\pm\) 1.8 & 8.3 \(\pm\) 2.2 \\
        AutoCompaction & 6.0 \(\pm\) 3.4 & 6.0 \(\pm\) 3.4 & 6.0 \(\pm\) 3.4 & 6.0 \(\pm\) 3.4 & 1.5 \(\pm\) 1.5 & 2.0 \(\pm\) 2.0 & 4.5 \(\pm\) 1.7 & 4.7 \(\pm\) 1.7 \\
        \midrule
        \textbf{Librarian} & \textbf{36.0 \(\pm\) 5.6} & \textbf{50.0 \(\pm\) 6.9} & \textbf{34.5 \(\pm\) 4.9} & \textbf{57.0 \(\pm\) 6.5} & \textbf{85.5 \(\pm\) 3.9} & \textbf{93.5 \(\pm\) 3.4} & \textbf{52.0 \(\pm\) 2.8} & \textbf{66.8 \(\pm\) 3.4} \\
        \bottomrule
    \end{tabular*}
    \endgroup
    \label{tab:bright-generalization}
\end{table*}

\noindent \textbf{(3) Transfer of Builder Agent to novel long-context document reasoning setting.}
To assess whether the builder agent learns a generalized anticipatory capability rather than simply overfitting to the DeepResearch training distribution, we evaluate \methodname{} zero-shot on the BRIGHT benchmark. Here, we study 150 questions spanning 3 domains with 50 per domain, with one filesystem per domain (also studying amortization). As detailed in Table \ref{tab:bright-generalization}, the RL-finetuned builder agent successfully transfers to highly out-of-distribution domains such as biology and math problem solving. \methodname{} achieves an average pass@1 accuracy of 0.52 and pass@3 accuracy of 0.67 across the 3 BRIGHT datasets, outperforming baselines. These results confirm that framing memory construction as an MDP enables the builder to learn fundamental, transferable heuristics for organizing information, which remain robust even when encountering novel, unseen documents.

The structural statistics on BRIGHT further show that RL substantially changes construction behavior. Across the 3 sub-domains, the trained builder produces richer hierarchical memories, averaging 53.4 active synthetic files organized into 45.4 clusters and 8.0 merged artifacts, compared with 4.8 files, 4.2 clusters, and 0.6 merges for the base builder. 

\begin{figure}[htbp]
    \centering
    \includegraphics[width=0.9\linewidth]{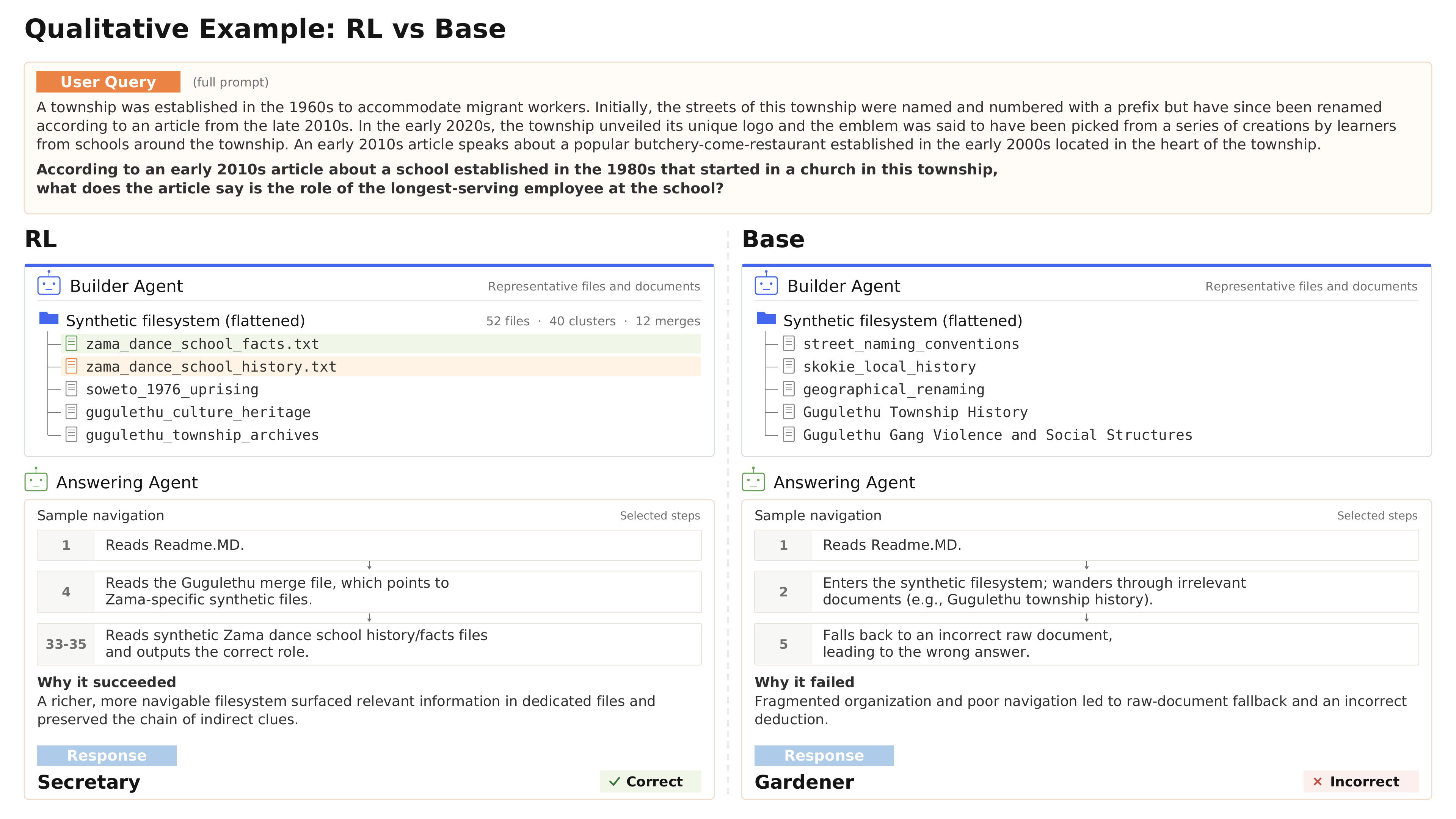}
    \caption{\footnotesize \textbf{Qualitative comparison of RL and base building.} On BrowseComp-Plus, the base model creates a fragmented filesystem that is difficult for the answering agent to navigate, whereas the RL-trained builder surfaces relevant information in dedicated files and produces a richer, more ergonomic organization.}
    \label{fig:rlvsbase}
\end{figure}
\noindent \textbf{(4) Analysis of \methodname{} and capability of anticipation.}
We next examine the filesystems that \methodname{} constructs and how the answering agent uses them. Figure~\ref{fig:mdp-browsecomp} illustrates that the trained builder combines broad topical coverage with hierarchical organization. Tables~\ref{tab:bright-stats} and~\ref{tab:browsecomp-stats} complement this view by distinguishing answering-agent retrieval (the synthetic files and tokens read) from builder structure (the active files, clusters, and merges it produces). Broad coverage expands the range of future queries the memory can support, while hierarchical organization makes fine-grained evidence easier to locate.

The first qualitative case examines robustness when anticipatory coverage is incomplete. In Figure~\ref{fig:recovery}, a BrowseComp+ query requires a detail omitted from the synthetic layer. The answering agent follows a provenance link to the relevant raw file and recovers the answer without enumerating the full unstructured corpus.

The second case isolates the changes that reinforcement learning makes in the constructed memory. Figure~\ref{fig:rlvsbase} compares the trained builder with the base policy on the same BrowseComp+ document set. The trained builder creates a broader, more coherent set of query-relevant artifacts, while the base policy leaves a fragmented organization that is harder for the answering agent to navigate.

\begin{AIbox}[Overview of Experimental Evaluation]
Our evaluation shows that \methodname{} improves long-context document reasoning in-domain, transfers to BRIGHT, amortizes construction cost across repeated queries, and learns memory structures that remain navigable when synthetic coverage is incomplete.
\end{AIbox}

\section{Related Work}




\textbf{Retrieval-Augmented Generation.} RAG grounds language model outputs in external evidence, mitigating the limitations of purely parametric memory \citep{Lewis2020RAG, Gao2023RAGSurvey, Li2025EnhancingRAG, Oche2025Systematic}. Recent innovations have evolved this pipeline into agentic systems capable of multi-step reasoning \citep{Xu2024ActiveRAG, Zhang2025TeaRAG, Singh2025AgenticRAGSurvey}. For instance, models can now predict upcoming content to trigger searches \citep{Jiang2023FLARE}, use reflection tokens for self-critique \citep{Asai2023SelfRAG, Yan2024CRAG}, or execute planned searches to navigate multi-hop questions \citep{PAR2RAG2026}. However, these approaches initiate complex, iterative retrieval only \emph{after} receiving a user query. In contrast, \methodname{} front-loads this computational burden by proactively synthesizing a structured, query-agnostic filesystem \emph{prior} to inference, amortizing the cost of corpus-level reasoning.

\textbf{Structured Indexes and Corpus-Level Organization.} Prior systems precompute corpus structure: RAPTOR builds hierarchical summaries \citep{Sarthi2024RAPTOR}; GraphRAG and related methods construct entity graphs and community summaries \citep{Edge2024GraphRAG, Li2024SubgraphRAG, Guo2025Empowering}; and HippoRAG builds graph-structured memory \citep{JimenezGutierrez2024HippoRAG}. Although valuable, these preprocessing methods rely heavily on rigid, rule-based extraction pipelines or predefined schemas. In contrast, \methodname{} make anticipatory memory construction fully agentic and optimizable. \methodname{} treats the structure itself as a learned decision, with a builder that chooses its own files and hierarchy and is rewarded by how well a later agent answers.

\textbf{Long-Term Memory for Language Agents.} Prior work explores endowing language models with persistent, stateful memory to support continuous operation. For instance, recent works propose to cache conversational histories \citep{Wang2023LongMem, Zhong2023MemoryBank, Uddin2026Benchmarking},  manage virtual context windows via memory tiers \citep{Packer2023MemGPT}, and unify short- and long-term memory directly within the agent's policy \citep{Yu2026AgeMem, Xu2025AMEM}. However, these approaches primarily focus on an agent retaining its own historical context. \methodname{} addresses a different setting, in which the memory is written over an external corpus ahead of time and judged by how well another agent answers from it.

\textbf{Compute Reallocation.} Our approach aligns with emerging literature on reallocating computation across time to improve test-time scaling \citep{Lin2025SleepTimeCompute, Wu2026ReasoningCache, yan-etal-2026-memory, yu2026memagentreshapinglongcontextllm,li2026machine}. Other approaches formalize test-time compute optimization as meta-reinforcement learning optimization to adaptively allocate tokens \citep{Qu2025MRT} or utilize self-curated test-time curricula to reinforce successful strategies \citep{Hubotter2025TTCRL}. To our knowledge, \methodname{} is the first to treat corpus organization as a policy optimized for downstream answers, training a builder to do once what an answering agent would otherwise repeat for every query. Our hyperparameters explicitly control how construction cost and query-time cost are weighed against answer accuracy.

\section{Discussion, Limitations, and Future Work}

We study the capability of \emph{anticipatory directed curation}, in which agents can proactively reorganize their context into a memory artifact optimized for future queries they encounter. To instantiate this capability, we rely on an asymmetry for agents today, where agents are trained to navigate a filesystem for a given task, yet do not pre-construct workspaces for future queries that they might encounter. Our novel framework, \methodname{}, introduces a builder agent that creates a synthetic filesystem that an answering agent can navigate. \methodname{} yields consistent improvements on BrowseComp-Plus and BRIGHT, outperforming existing retrieval methods on both in-domain and out-of-distribution evaluations.

Our evaluation is limited to question answering over static corpora, and maintaining anticipatory memories as documents change remains an open problem. The builder agent also operates over a large file context, which may limit scalability, though pairing Librarian Models with clustering or retrieval could extend the framework to larger domains. Since no finite set of synthesized artifacts can anticipate every query, compression may discard details needed for needle-in-a-haystack tasks; we preserve reference links for fallback access, but deciding when to bypass the synthetic filesystem remains a challenge. A natural extension is a filesystem that updates from its own traffic, treating each answered query as evidence of what the next reader will need. Future work could also let the builder compose atomic filesystem commands, of which its current four actions are special cases.
\vspace{-0.2cm}
\section{Acknowledgements}
\vspace{-0.2cm}

We would like to thank Apurva Gandhi, Jack Hao Bai, Violet Xiang, Rulin Shao and others in the Stanford IRIS Lab and CMU AIRE Lab for discussions. We thank Simon Guo, Rafael Rafailov, and Charles Frye for their support via the Tinker Research Program and the Modal Academic Research Program. This work was supported by the Schmidt Sciences AI2050, NSF GRFP and an Amazon gift.
\clearpage
\bibliographystyle{iclr2024_conference}
\bibliography{iclr2024_conference}
\clearpage
\appendix
\section{Appendix}
\label{sec:appendix}
\subsection{Algorithm Overview of \methodname{}}
\begin{algorithm}[H]
\caption{RL Training of \methodname{}}
\label{alg:memory-construction}
\scriptsize
\begin{algorithmic}[1]
    \Require Documents $\mathcal{D}$, query distribution $\mathcal{Q}$, builder $\pi_{\text{build}}$, answerer $\pi_{\text{ans}}$, horizon $H$
    \Ensure Optimized builder policy $\pi_{\text{build}}$

    \For{$k = 1,\ldots,K$}
        \State Sample hidden reward-query batch $\mathcal{B}_k \sim \mathcal{Q}$
        \State Initialize $\mathcal{F}_0 \gets (\mathcal{D}, \emptyset)$, $\tau \gets \emptyset$

        \For{$t = 1,\ldots,H$}
            \State Sample edit $\mathcal{A}_t \sim \pi_{\mathrm{build}}(\cdot \mid \mathcal{F}_{t-1})$
            \State Initialize $\widehat{r}_t \gets 0$
            \If{$\mathcal{A}_t = \textsc{Stop}$}
                \State $\mathcal{F}_t \gets \mathcal{F}_{t-1}$
                \State Append $(\mathcal{F}_{t-1}, \mathcal{A}_t, \widehat{r}_t, \mathcal{F}_t)$ to $\tau$
                \State \textbf{break}
            \EndIf
            \State Apply edit: $\mathcal{F}_t \gets \mathcal{A}_t(\mathcal{F}_{t-1})$
            \If{$t$ is an answerability-probe step}
                \State Estimate $\widehat{S}(\mathcal{F}_t)$ on $\mathcal{B}_k$
                \State Assign answerability-delta reward $\widehat{r}_t$
            \EndIf
            \State Append $(\mathcal{F}_{t-1}, \mathcal{A}_t, \widehat{r}_t, \mathcal{F}_t)$ to $\tau$
        \EndFor
        \State Evaluate terminal answer score on $\mathcal{B}_k$
        \State Add the terminal score to the rollout reward

        \State Update $\pi_{\text{build}}$ with GRPO on rollout $\tau$
    \EndFor

    \State \Return $\pi_{\text{build}}$
\end{algorithmic}
\end{algorithm}

\subsection{More details on experimental setup}
\paragraph{BrowseComp-Plus.} BrowseComp-Plus~\citep{chen2025browsecomp} pairs document collections with questions that require locating and combining information distributed across those collections. We train the filesystem builder on 680 corpus-level tasks and reserve 50 previously unseen corpus-level tasks, for evaluation. We use a mixture of evidence, gold, and negative docs for each representative task, following the composition in the original dataset. During construction, the builder receives only the documents and never observes the evaluation questions. The completed filesystem is then reused for every question associated with its corpus. We report failure-inclusive binary semantic answer correctness against the fixed reference answers, so unsuccessful constructions and failed generations remain in the denominator. In addition, for the Base-versus-RL comparison, both builders receive identical construction limits, and we report realized construction tokens separately because equal limits do not imply equal consumption.

\paragraph{BRIGHT-derived question answering.} BRIGHT spans technical, scientific, mathematical, and programming domains with substantially different document and query distributions~\citep{su2025brightrealisticchallengingbenchmark}. We convert its retrieval instances into a generation task by associating each query with a fixed reference answer and scoring generated responses for semantic correctness. For the primary matched-baseline evaluation, we use 150 questions spanning three domains, with 50 questions each from AOPS, Biology, and Stack Overflow. To test cross-benchmark transfer, we apply the BrowseComp-trained Librarian zero-shot: it constructs one reusable filesystem per domain without observing any evaluation question, and this single shared filesystem supports all questions in that domain. We compare the Librarian against matched retrieval baselines, generating four independent responses per question and reporting both mean per-response accuracy, which estimates pass@1, and unbiased pass@3 estimated from the four generated responses, which we compute using the numerically stable estimator from Codex~\citep{chen2021evaluatinglargelanguagemodels}.

\subsection{Additional Results}
First, we list statistics on the number of synthetic files and tokens created by the builder agent. We also include details about the breadth and depth of the constructed filesystems from the builder.

\begin{table}[htbp]
    \centering
    \scriptsize
    \setlength{\tabcolsep}{4pt}
    \renewcommand{\arraystretch}{1.08}
    \caption{
        \footnotesize
        \textbf{BRIGHT statistics by domain.}
        We report answerer retrieval behavior and final filesystem-structure statistics for BRIGHT. Syn. Files, Syn. Tok., and Tot. Tok. measure files and tokens read by the answering agent, while Act. Files, Clusters, and Merges summarize the active synthetic filesystem produced by the builder.
    }
    \begin{tabular}{llrrrrrr}
        \toprule
        \textbf{Domain} & \textbf{Model} & \textbf{Syn. Files} & \textbf{Syn. Tok.} & \textbf{Tot. Tok.} & \textbf{Act. Files} & \textbf{Clusters} & \textbf{Merges} \\
        \midrule
        Bio. & Base & 2.7 & 1064 & 1189 & 5.4 & 4.9 & 0.5 \\
         & RL & 30.4 & 9883 & 10091 & 54.0 & 45.8 & 8.2 \\
        StackOv. & Base & 3.3 & 1075 & 1250 & 4.5 & 4.0 & 0.5 \\
         & RL & 32.0 & 9246 & 9511 & 49.5 & 40.6 & 8.8 \\
        AoPS & Base & 3.9 & 879 & 1023 & 4.5 & 3.8 & 0.7 \\
         & RL & 36.2 & 7309 & 7544 & 56.7 & 49.7 & 7.0 \\
        \midrule
        \textbf{Macro avg.} & Base & 3.3 & 1006 & 1154 & 4.8 & 4.2 & 0.6 \\
         & \textbf{RL} & 32.9 & 8813 & 9049 & 53.4 & 45.4 & 8.0 \\
        \bottomrule
    \end{tabular}%
    \label{tab:bright-stats}
\end{table}

\begin{table}[htbp]
    \centering
    \scriptsize
    \setlength{\tabcolsep}{4pt}
    \renewcommand{\arraystretch}{1.08}
    \caption{
        \footnotesize
        \textbf{BrowseComp+ statistics.}
        We report the same statistics on the amortized BrowseComp+ evaluation. Syn. Files, Syn. Tok., and Tot. Tok. again measure answering-agent retrieval behavior, while Act. Files, Clusters, and Merges summarize the active synthetic filesystem produced by the builder.
    }
    \begin{tabular}{llrrrrrr}
        \toprule
        \textbf{Dataset} & \textbf{Model} & \textbf{Syn. Files} & \textbf{Syn. Tok.} & \textbf{Tot. Tok.} & \textbf{Act. Files} & \textbf{Clusters} & \textbf{Merges} \\
        \midrule
        BrowseComp+ & Base & 2.7 & 1069 & 1235 & 3.2 & 2.7 & 0.6 \\
         & RL & 26.9 & 9040 & 9309 & 42.2 & 34.4 & 7.8 \\
        \bottomrule
    \end{tabular}%
    \label{tab:browsecomp-stats}
\end{table}

Finally, we include additional bright transfer results comparing \methodname{} against the base model in Table~\ref{tab:bright-additional-domains}. We see consistent improvement across the nine additional domains within the Bright dataset, using $n=20$ questions per domain.

\begin{table*}[htbp]
    \centering
    \caption{
        \footnotesize
        \textbf{Comparison with the Base Model on Additional BRIGHT Domains.}
        \methodname{} outperforms the base model across all nine
        additional domains. Results are reported as percentages.
    }
    \resizebox{\textwidth}{!}{%
    \begin{tabular}{l*{9}{c}}
    \toprule
    Method & \shortstack{Earth\\Science} & Economics & Psychology & Robotics & \shortstack{Sustainable\\Living} & LeetCode & Pony & \shortstack{TheoremQA\\Questions} & \shortstack{TheoremQA\\Theorems} \\
    \midrule
    Base Model & 26.3\% & 16.3\% & 32.5\% & 23.7\% & 12.5\% & 28.7\% & 48.7\% & 35.0\% & 38.8\% \\
    \textbf{Librarian} & \textbf{73.8\%} & \textbf{72.5\%} & \textbf{72.5\%} & \textbf{55.0\%} & \textbf{43.8\%} & \textbf{56.2\%} & \textbf{72.5\%} & \textbf{63.7\%} & \textbf{56.2\%} \\
    \bottomrule
    \end{tabular}%
    }
    \label{tab:bright-additional-domains}
\end{table*}

\subsection{BrowseComp-Plus Reuse and Break-Even Analysis}
\label{app:browsecomp-break-even}

\begin{figure*}[htbp]
\centering
\includegraphics[width=0.48\textwidth]{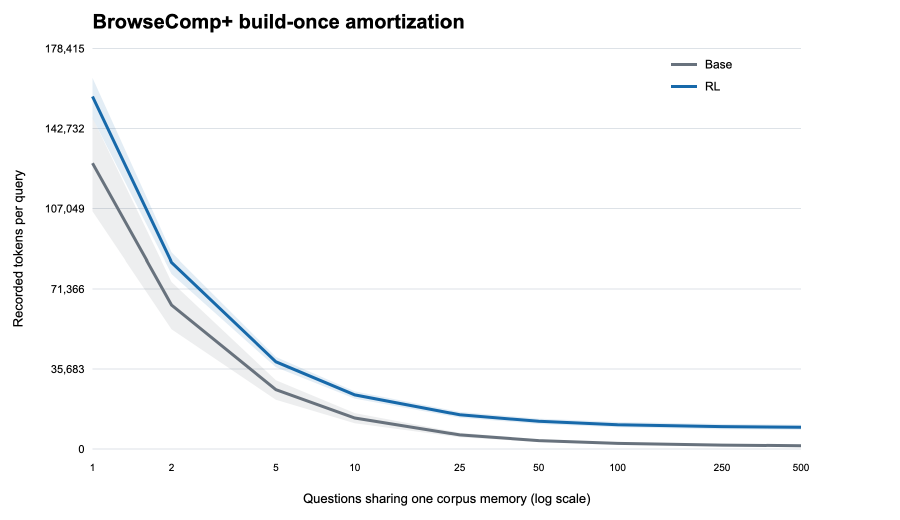}
\hfill
\includegraphics[width=0.48\textwidth]{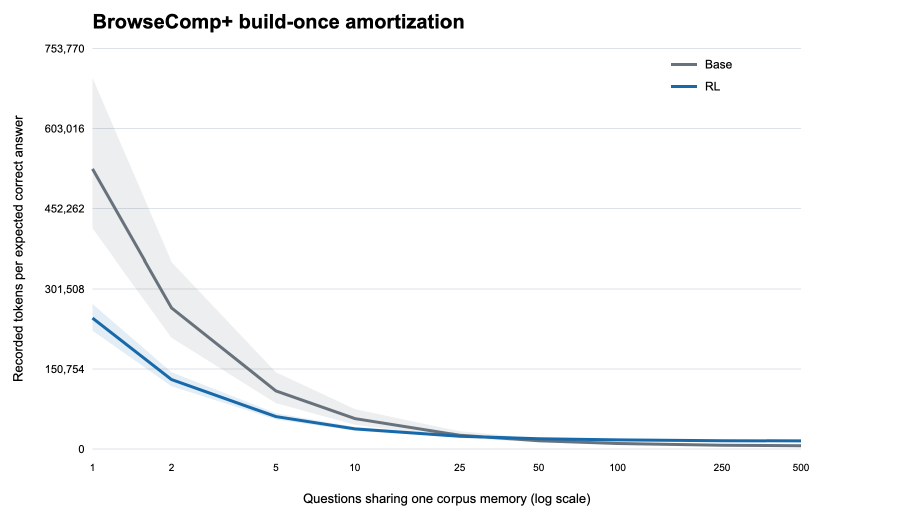}
\caption{\textbf{BrowseComp-Plus reuse curves.} Left: recorded construction and navigation tokens per query. Right: recorded tokens per expected correct answer after normalization by observed full-denominator accuracy. Shaded regions show paired corpus-bootstrap 95\% confidence intervals.}
\label{fig:browsecomp-break-even}
\end{figure*}

We compute reuse curves directly from the frozen BrowseComp-Plus rollout logs. For method $m$ and $q$ questions per corpus, recorded tokens per query are
\[
T_m(q)=\frac{B_m}{q}+A_m,
\]
where $B_m$ is the observed construction-token cost per corpus and $A_m$ is the observed answer-time workspace-token cost per question. To account for differences in answer quality, we additionally report $T_m(q)/p_m$, where $p_m$ is full-denominator accuracy.

At ten questions per corpus, RL uses 33.8\% fewer recorded tokens per expected correct answer than the base builder (RL/Base ratio $0.662$, paired-bootstrap 95\% CI $[0.504,0.836]$). The point crossover occurs at 31 questions per corpus, with a paired-bootstrap interval of 18--54 questions. Because the RL answerer explores a richer workspace, its raw recorded tokens per query remain higher; we therefore report raw and accuracy-normalized curves together rather than treating token count independently of answer quality.

\subsection{Filesystem-structure ablation.}
To isolate the effect of filesystem organization, we evaluate the 49 corpora
with completed filesystem snapshots, comprising 435 questions from the expanded amortization set. Unlike the original one-question-per-corpus BrowseComp+ evaluation reported in Figure~3, this analysis uses the expanded questions and a fresh matched answering run. Accordingly, comparisons are made within this run across the original, flattened, and shuffled conditions. We apply two content-preserving interventions to the same memories. Flattening preserves comparable answer accuracy but increases navigation cost, while shuffling semantic placement reduces accuracy without reducing navigation cost.

\begin{table*}[htbp]
\centering
\small
\caption{\footnotesize \textbf{Filesystem-structure ablation on BrowseComp+.}
Accuracy and answering-agent navigation-token cost are evaluated on 435
expanded amortization questions across the 49 corpora with completed filesystem
snapshots. All three conditions use the same matched answering protocol.
Standard errors use a corpus-cluster bootstrap.}
\label{tab:hierarchy-ablation}
\begin{tabular}{lrr}
\toprule
Memory organization & Accuracy (\%) & Navigation tokens \\
\midrule
Original  & $55.86 \pm 2.58$ & $3{,}635 \pm 216$ \\
Flattened & $57.70 \pm 2.81$ & $5{,}497 \pm 356$ \\
Shuffled  & $51.72 \pm 2.87$ & $3{,}612 \pm 221$ \\
\bottomrule
\end{tabular}
\end{table*}

\subsection{Qualitative Example of Recovery Using Reference Information}

\begin{figure}[htbp]
    \centering
    \includegraphics[width=\linewidth]{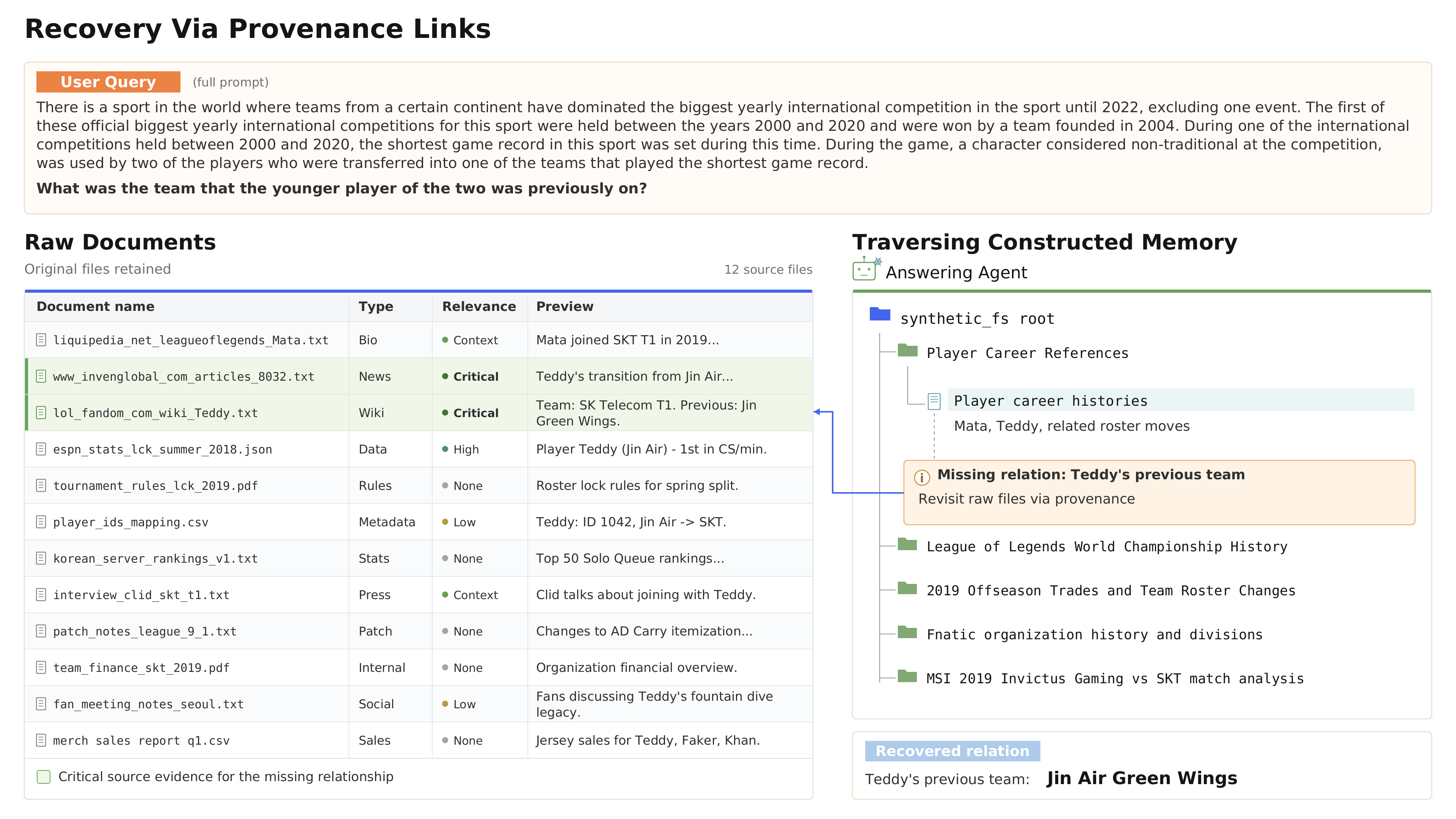}
    \caption{\footnotesize \textbf{Recovery on BrowseComp-Plus using provenance information.} When the synthetic filesystem omitted a required detail, the answering agent followed provenance links to the relevant raw source and recovered the correct answer.}
    \label{fig:recovery}
\end{figure}

We also showcase a qualitative example of an agent recovering using Reference Information, when the synthesized documents didn't have the complete information to answer a question in Figure~\ref{fig:recovery}.

\subsection{\methodname{} Training and Evaluation Configuration}
\label{app:hyperparam}

For the primary BrowseComp-Plus experiment, we optimize a Qwen3.5-4B builder with GRPO on the 680-task training split. We use LoRA rank 32, a learning rate of $4\times10^{-5}$, batch size 8, rollout-group size 4, and a maximum construction horizon of 32 builder turns. Builder generations are limited to 1,024 tokens and complete construction trajectories to 60,000 tokens. The \textsc{CreateFile} and \textsc{MergeFiles} operations use a fixed Qwen3.5-35B-A3B content executor, and long builder histories are compacted by a fixed Gemini 3.1 Flash Lite model. The answering agent and evaluator also remain fixed throughout builder optimization.

The training reward measures the downstream usefulness of the evolving filesystem. We probe answerability every eight builder turns, with at most four probes per construction trajectory, and assign reward according to the change in answerability between consecutive probes. A terminal reward evaluates the final filesystem using downstream answer correctness. We use $\gamma=1$ and no explicit filesystem-structure reward, navigation-cost reward, or construction-cost shaping in the primary run ($\alpha=0$). Filesystem maturity, organization, and navigation measurements reported in the analysis are therefore diagnostic statistics rather than direct components of the optimization objective.

The matched BRIGHT evaluation uses temperature $0.7$ and top-$p=1$ for answer sampling, while single-answer evaluations use deterministic decoding. Within each comparison, all methods use the same source corpus, answering model, response count, answer-generation budget, and scoring procedure. Evaluation questions are never shown to the builder during construction or used to select the reported checkpoint.

\subsection{Tool Definitions for Filesystem Action Space}
\label{app:tool-use}

The builder agent interacts with the filesystem by delegating a limited set of discrete operations to specialized tool-calling agents. These tools are made available to the builder LLM through structured JSON schemas and can be invoked as needed. We follow the Qwen Tool Calling Format: \url{https://qwen.readthedocs.io/en/latest/framework/function_call.html}.

The CREATEFILE tool creates a new synthetic file by consolidating information from selected raw documents and/or existing synthetic files. The resulting file is added to the hierarchy and includes provenance links to the sources used to generate it. The MERGEFILES tool combines related synthetic files into a higher-level artifact, such as a summary, index, or directory-like file. This operation introduces hierarchical structure while preserving links to the underlying child files, which become nested under the new summary file.

The DELETE tool removes redundant or unhelpful synthetic files from the active, answer-facing filesystem. It never deletes raw documents in $\mathcal{D}$, ensuring that the original corpus remains available as a fallback. Finally, the STOP tool ends the memory-construction process and returns the current filesystem $\mathcal{F}_t$ as the final memory artifact.

We use the agent’s function-calling capability to invoke these specialized subagents, each of which implements one of the individual tools. 

\subsection{\methodname{} and Baseline Prompts}
\label{app:prompts}
To initialize the builder agent, we utilize a detailed system prompt that establishes its task and action space constraints: 

\begin{AIBox}[Prompt]
\begin{Verbatim}[
  breaklines=true,
  breakanywhere=true,
  fontsize=\footnotesize,
  baselinestretch=0.96,
  obeytabs=true,
  tabsize=2
]
You are building a synthetic filesystem for a later answering agent.

Use the provided filesystem tools to inspect files under:
- raw_docs/
- synthetic_fs/
- history/

Your job is to leave behind a final active synthetic filesystem that helps later agents answer as many plausible future questions about this corpus as possible.
Do not optimize for one guessed target question; build broad, reusable knowledge structure over the corpus.
The later answering agent will start with root-level synthetic entries, browse deeper synthetic files if needed, and use raw_docs/ only as fallback.
\end{Verbatim}
\end{AIBox}

Here is the prompt for the answering agent.

\begin{AIBox}[Prompt]
\begin{Verbatim}[
  breaklines=true,
  breakanywhere=true,
  fontsize=\footnotesize,
  baselinestretch=0.96,
  obeytabs=true,
  tabsize=2
]
You are the answering agent for a read-only workspace.

You have filesystem tools over a final workspace that contains:
- README.md explaining the workspace conventions and navigation only
- root-level synthetic entries that are already loaded into context as the primary entry points
- nested synthetic folders and summary files for more detail

Use this order:
1. start from the already-loaded README.md and root-level synthetic entries
2. browse deeper synthetic files if needed

Do not guess. If the synthetic files you have read so far do not contain enough evidence to uniquely identify the answer, continue exploring the synthetic filesystem before answering. README.md only explains the workspace and is not sufficient evidence by itself. Answer only when the synthetic workspace contains enough evidence.

At each step, either:
- return exactly one compact JSON action:
  {"action":"list_files","path":"."}
  {"action":"read_file","path":"some/path.txt","start_line":1,"num_lines":80}
- or return exactly one non-empty line:
  FINAL ANSWER: <final answer>

Do not include explanation outside the JSON action or final answer line.
\end{Verbatim}
\end{AIBox}

We use the following prompt for Judging with the Gemini 3.1 Flash Lite model as a judge for our method and baselines.
\begin{AIBox}[Prompt]
\begin{Verbatim}[
  breaklines=true,
  breakanywhere=true,
  fontsize=\footnotesize,
  baselinestretch=0.96,
  obeytabs=true,
  tabsize=2
]
You are given a question, sampled response, and gold answer.
Decide whether the sampled response is semantically equivalent to the gold answer.
Minor formatting, capitalization, punctuation, and unit-format differences are acceptable
when the answer means the same thing.

Return only valid JSON with this shape:
{{"correctness": 0 or 1}}

QUESTION:
{question}

SAMPLED RESPONSE:
{response}

GOLD ANSWER:
{gold_answer}
\end{Verbatim}
\end{AIBox}

For baselines, the common answering agent prompt was:
\begin{AIBox}[Prompt]
\begin{Verbatim}[
  breaklines=true,
  breakanywhere=true,
  fontsize=\footnotesize,
  baselinestretch=0.96,
  obeytabs=true,
  tabsize=2
]
You are a grounded question-answering assistant.

Answer the question using the retrieved documents below.

Instructions:
1. Treat the retrieved documents as a source of truth.
2. Some documents may be irrelevant, incomplete, duplicated, or conflicting.
3. Ignore any document content that does not directly help answer the question.
4. Do not use outside knowledge or make assumptions.
5. If documents conflict, use the answer best supported by the most direct evidence.
6. Return only the final short answer. Do not include explanation, citations, markdown, or extra text.

QUESTION:
{question}

RETRIEVED DOCUMENTS:
{documents}

FINAL ANSWER:
\end{Verbatim}
\end{AIBox}

We also list the prompts for the GraphRag baseline, following the microsoft graph rag system: \url{https://github.com/microsoft/graphrag}. 

Here is the graph extraction prompt.
\begin{AIBox}[Prompt]
\begin{Verbatim}[
  breaklines=true,
  breakanywhere=true,
  fontsize=\footnotesize,
  baselinestretch=0.96,
  obeytabs=true,
  tabsize=2
]
You are building a knowledge graph for retrieval-augmented question answering.

Goal:
Given one text unit, extract the salient entities and relationships needed to build a reusable corpus-level
knowledge graph for retrieval-augmented question answering.
Keep specific entities, bridge entities, dates, quantities, events, works, and facts that could support
future multi-hop questions over this corpus.

Entity types to consider:
{entity_types}

Output format:
Use "{record_delimiter}" between records and finish with "{completion_delimiter}".
For each entity:
("entity"<|>ENTITY_NAME<|>ENTITY_TYPE<|>ENTITY_DESCRIPTION)
For each relationship:
("relationship"<|>SOURCE_ENTITY_NAME<|>TARGET_ENTITY_NAME<|>RELATIONSHIP_DESCRIPTION<|>RELATIONSHIP_STRENGTH)

Rules:
- Use canonical, specific entity names.
- Descriptions must be concise and grounded in the text unit.
- Relationship strength is a number from 1 to 10.
- Do not infer facts beyond the text unit.
- If there are no useful records, output only "{completion_delimiter}".

Text unit metadata:
path: {path}
text_unit_id: {text_unit_id}

Text unit:
{text}
\end{Verbatim}
\end{AIBox}

Here is the summarization description prompt:
\begin{AIBox}[Prompt]
\begin{Verbatim}[
  breaklines=true,
  breakanywhere=true,
  fontsize=\footnotesize,
  baselinestretch=0.96,
  obeytabs=true,
  tabsize=2
]
You are reconciling extracted knowledge graph descriptions.

Given one entity or relationship name and several source descriptions, write one concise, faithful summary.
Preserve specific names, dates, quantities, and qualifications that may matter for question answering.
Do not add facts that are not present in the descriptions.

Name:
{name}

Descriptions:
{descriptions}

Return only the merged summary.
{text}
\end{Verbatim}
\end{AIBox}

Here is the community report prompt:
\begin{AIBox}[Prompt]
\begin{Verbatim}[
  breaklines=true,
  breakanywhere=true,
  fontsize=\footnotesize,
  baselinestretch=0.96,
  obeytabs=true,
  tabsize=2
]
You are writing a GraphRAG community report for retrieval-augmented question answering.

Use only the provided entity and relationship tables. The report should summarize what this community contains
so it can be reused for later retrieval-augmented question answering.

Return only valid JSON with this shape:
{{
  "title": "short community title",
  "summary": "one paragraph summary grounded in the tables",
  "rating": 0-10,
  "findings": [
    {{"summary": "specific finding", "explanation": "brief explanation with entity or relationship ids"}}
  ]
}}

Maximum report words: {max_report_words}

Community tables:
{community_context}
\end{Verbatim}
\end{AIBox}

For autocompaction, we use the following prompt for summarization:

\begin{AIBox}[Prompt]
\begin{Verbatim}[
  breaklines=true,
  breakanywhere=true,
  fontsize=\footnotesize,
  baselinestretch=0.96,
  obeytabs=true,
  tabsize=2
]
Compress a document stream into query-independent notes for later QA.

Rules:
- The future question is hidden. Do not tailor the notes to any query.
- Keep only grounded facts: names, dates, numbers, identifiers, titles, events, relationships, claims, and caveats.
- Prefer dense bullet notes or compact clauses. Remove filler, repeated wording, and low-value prose.
- If facts conflict, keep the conflict and source path instead of resolving it.
- Use at most {summary_max_tokens} tokens.
- Return only the updated notes. Do not include reasoning, analysis, preambles, or markdown fences.

CURRENT SUMMARY:
{current_summary}

NEXT DOCUMENT:
path: {path}
stream_position: {stream_position}
fragment: {fragment_index}/{num_fragments}
content:
{document_text}
\end{Verbatim}
\end{AIBox}

For agentic rag, we use the following prompt, based on the smolagents implementation: \url{https://huggingface.co/docs/smolagents/en/examples/rag}.
\begin{AIBox}[Prompt]
\begin{Verbatim}[
  breaklines=true,
  breakanywhere=true,
  fontsize=\footnotesize,
  baselinestretch=0.96,
  obeytabs=true,
  tabsize=2
]

Use the retrieval history below to decide the next tool call.

Available tools:
1. search: retrieve documents for one new query.
2. final_results: stop and choose the final document paths.

Return only valid JSON, with one of these shapes:
{{"thought": "brief reason", "action": "search", "query": "new dense-retrieval query"}}
{{"thought": "brief reason", "action": "final_results", "document_paths": ["path/a.txt", "path/b.txt"]}}

Do not answer the question. Search queries should be specific, diverse, and useful for finding missing evidence.
If agent_searches_remaining is 0, return final_results.

PAYLOAD:
\end{Verbatim}
\end{AIBox}

\subsection{Training Curve for \methodname{}}

\begin{figure}[htbp]
    \centering
    \includegraphics[width=\linewidth]{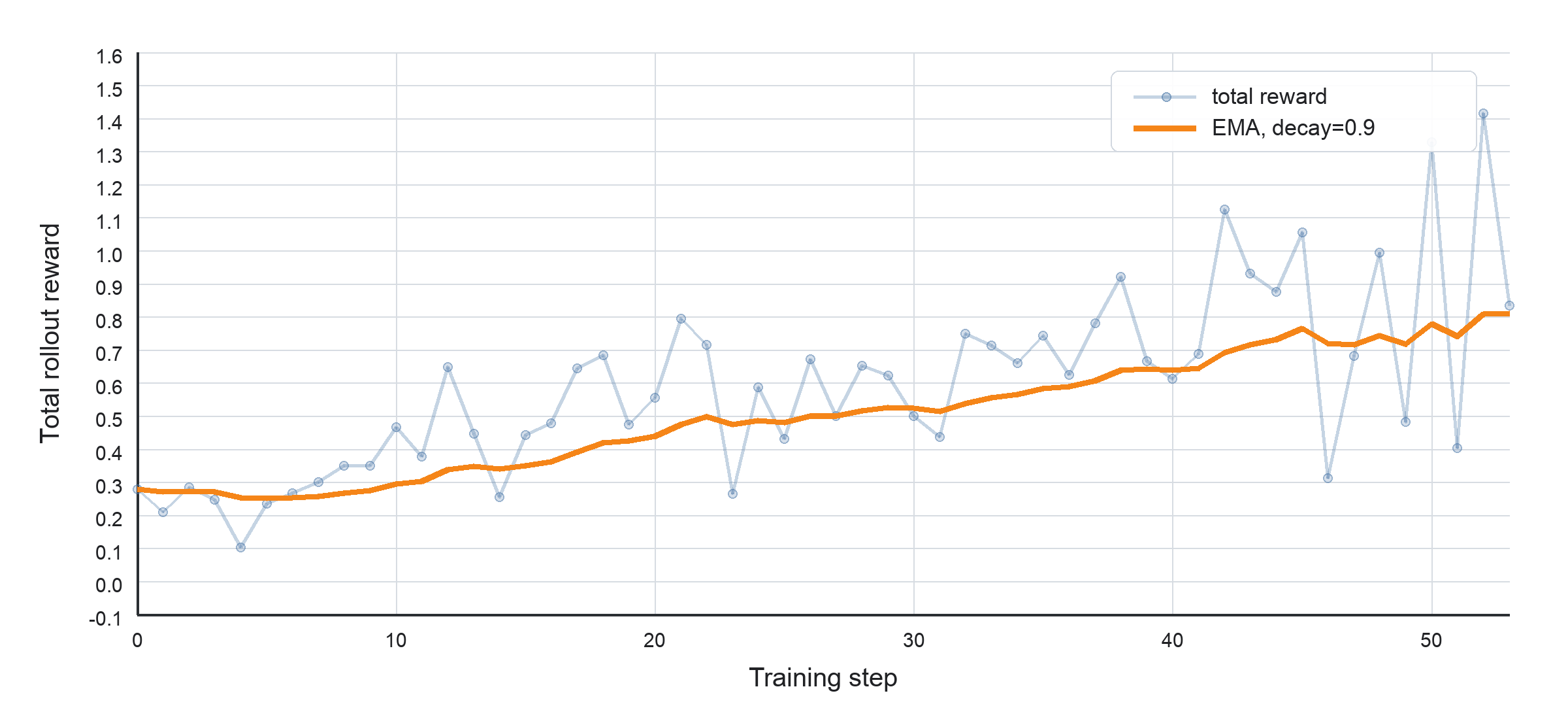}
    \caption{\footnotesize \textbf{Training curve for the BrowseComp-Plus builder.} Aggregate rollout reward during GRPO training, smoothed with an exponential moving average.}
\label{fig:training_curve}
\end{figure}

Figure~\ref{fig:training_curve} reports the aggregate rollout reward during GRPO optimization on the BrowseComp-Plus training split. Reward increases over training as the builder learns construction decisions that produce more answerable downstream memories.

\subsection{License for Artifacts}
The code, training scripts, and evaluation framework for \methodname{} will be released under the MIT License. Synthesized filesystem artifacts, logs, and evaluation results generated during our experiments will be released under the Creative Commons Attribution 4.0 International (CC BY 4.0) license. The models used as answering agents, namely Qwen3.5-4B and Gemini 3.1 Flash Lite, are subject to their respective licensing agreements (e.g., the Tongyi Qianwen LICENSE agreement and Google's terms of service). BrowseComp-Plus, BRIGHT, and DeepDive datasets are subject to their creators' respective data licensing terms.

\end{document}